\documentclass[conference]{IEEEtran}

\makeatletter

\def\ps@IEEEtitlepagestyle{%
  \def\@oddfoot{\mycopyrightnotice}%
  \def\@evenfoot{}%
}
\def\mycopyrightnotice{%
  {\footnotesize XXX-X-XXXX-XXXX-X/XX/\$XX.00~\copyright~20XX IEEE\hfill}
  \gdef\mycopyrightnotice{}
}

\usepackage[caption=false,font=footnotesize]{subfig}

\usepackage{blindtext}
 \usepackage{url}
\usepackage{hyperref}   
\usepackage{breakurl}
\usepackage{eso-pic}
\IEEEoverridecommandlockouts
\usepackage{cite}
\usepackage{amsmath,amssymb,amsfonts}
\usepackage{algorithmic}
\usepackage{graphicx}
\usepackage{textcomp}
\usepackage{xcolor}
\def\BibTeX{{\rm B\kern-.05em{\sc i\kern-.025em b}\kern-.08em
    T\kern-.1667em\lower.7ex\hbox{E}\kern-.125emX}}
    
\usepackage{eso-pic}
\newcommand\AtPageUpperMyright[1]{\AtPageUpperLeft{%
 \put(\LenToUnit{0.17\paperwidth},\LenToUnit{-2cm}){%
     \parbox{0.9\textwidth}{\raggedleft\fontsize{8}{11}\selectfont #1}}%
 }}%
\newcommand{\conf}[1]{%
\AddToShipoutPictureBG*{%
\AtPageUpperMyright{#1}
}
}    

\usepackage{placeins}

\newtheorem{example}{Example}[section]

\begin{document}
\title{\vspace*{1cm} Partitioning the Sample Space for a More Precise Shannon Entropy Estimation\\
\thanks{This study was financed in part by the Coordenação de Aperfeiçoamento 
de Pessoal de Nível Superior – Brasil (CAPES) – Finance Code 001}
}

\author{\IEEEauthorblockN{1\textsuperscript{st} Gabriel F.A. Bastos}
\IEEEauthorblockA{\textit{Electrical Engineering Department} \\
\textit{Federal University of Sergipe}\\
São Cristóvão, Brazil \\
gabrielfab210102@gmail.com}
\and
\IEEEauthorblockN{2\textsuperscript{nd} Jugurta Montalvão}
\IEEEauthorblockA{\textit{Electrical Engineering Department} \\
\textit{Federal University of Sergipe}\\
São Cristóvão, Brazil \\
jmontalvao@academico.ufs.br}
}

\maketitle
\conf{\textit{  Proc. of International Conference on Artificial Intelligence, Computer, Data Sciences and Applications (ACDSA 2026) \\ 
5-7 February 2026, Boracay-Philippines}}
\begin{abstract}
Reliable data-driven estimation of Shannon entropy from small data sets, where the number of examples is potentially smaller than the number of possible outcomes, is a critical matter in several applications. In this paper, we introduce a discrete entropy estimator, where we use the decomposability property in combination with estimations of the missing mass and the number of unseen outcomes to compensate for the negative bias induced by them. Experimental results show that the proposed method outperforms some classical estimators in undersampled regimes, and performs comparably with some well-established state-of-the-art estimators. 
\end{abstract}


\begin{IEEEkeywords}
entropy, decomposability, unseen, missing mass
\end{IEEEkeywords}

\section{Introduction}
Let $X$ be a discrete random variable supported on $\{1, \ldots, S\}$ with distribution $P = (p_1, \ldots, p_S)$, where $p_i = Pr(X=i)$. The Shannon entropy of the distribution $P$ is given by \begin{equation}\label{eq: def}
    H(P) = -\sum \limits _{i=1}^{S} p_i\log(p_i),
\end{equation} with the convention that $-p\log(p)=0$, for $p=0$. When it is convenient to think of the entropy as an attribute of the random variable, instead of its distribution, we shall write $H(P)$ as $H(X)$. Here, the logarithmic base defines the unit of entropy. For example, for base 2, the entropy is expressed in bits of information, while for base $e$, the entropy is expressed in nats.  

This statistical measure was introduced by Shannon in the context of the study of the fundamental limits of data compression and transmission over a communication channel \cite{shannon1948}. From a more general perspective, the entropy quantifies the amount of uncertainty involved in a random experiment, and quickly found uses in various other application fields, beyond communications, such as linguistics \cite{arora2022}, physiological signal analysis \cite{kafantaris2022}, machine learning \cite{mao2023} and ecology \cite{harte2014}.

Due to its broad range of applications, determining the entropy of a given distribution --- or of a random source --- is a fundamental matter. However, in practical situations, very rarely is the parameter $P$ known, and all the information available is a set of $N$ realizations from the random variable, which we denote by $X^N = (x_1, \ldots, x_N)$. Therefore, the problem of estimating the entropy is formulated as the construction of a function that takes as input a data set and outputs an approximate estimate of the entropy of the distribution that generated the input data set. That is, the goal is to obtain $\hat{H}$ such that $\hat{H}(X^N) \approx H(P)$.

Given that the entropy is a functional of a probability distribution, a natural approach to its estimation is to first obtain empirical estimates of the probabilities, given by $\hat{p}_i = n_i/N$, where $n_i$ is the number of times the symbol $i$ was observed in $X^N$, and then replace these in (\ref{eq: def}). Indeed, this estimator, known as plug in, is the simplest and most widely used one. However, it is well known that the plug in estimation is negatively biased when the sample size $N$ is small, potentially smaller than the support size $S$. On the other hand, applications where the sample size is smaller than the support size are increasingly more common. For instance, in the linguistic domain it is common to work with corpora where some words were not observed. In the neuroscience domain, the enormous amount of possible neural spike train patterns easily exceeds any amount of samples that is feasible to collect \cite{luczak2024}.   

Given that, the problem of developing estimators that achieves consistent results in small sample regimes has attracted the attention of researchers from different backgrounds over the last several decades. Early attempts include standard statistical bias-reduction techniques, such as the well known Miller-Madow estimator \cite{miller1955}. Bayesian regularization of the frequency counts using Dirichlet priors with given parameters may also lead to improvements in the empirical estimations, such as in \cite{schurmann1996}. In \cite{nemenman2001}, the authors avoid overrelying on a single parameter choice by using a Dirichlet mixture prior with infinite components. Still in the Bayesian framework, in \cite{hausser2009}, the authors propose an optimal estimation of the Dirichlet parameter in a data-driven manner, a technique named Shrinkage.  In \cite{chao2003}, the Chao-Shen estimator combines the Horvitz-Thompson estimator with sample coverage-corrected probabilities. To this day, the Chao-Shen estimator is widely regarded as one of the best available estimators, as shown in multiple studies such as in \cite{arora2022}. Other estimators use polynomial smoothing and linear programming to compensate the estimation bias \cite{wu2016, valiant2017, hao2018}.

In this paper, we apply the decomposability property to partition the support according to the frequency of occurrence of each symbol and use a different strategy to estimate the contribution of each part. This perspective enables us to directly tackle the main source of estimation bias in the data scarcity regime, which is the presence of unseen symbols in a sample. With that, we propose a simple and interpretable entropy estimator that performs comparably with some of the best ones available in the literature. 

This paper is organized as follows. In Section \ref{sec: empirical estimations}, we illustrate the main issues in empirical estimation. In Section \ref{sec: unseen}, we present the tools that allow us to directly tackle these issues. The proposed estimator is presented in Section \ref{sec: proposed} and experimentally evaluated in Section \ref{sec: experiments}. Section \ref{sec: conclusions} offers concluding remarks and outlines directions for future works. 

\section{What are the main source of errors in the empirical estimations?} \label{sec: empirical estimations}

In this Section, we illustrate the argument that the unseen symbols are the main source of estimation bias in plug in estimation of entropy. This is due to the fact that, in small samples, rare symbols are expected to not even be observed, and these symbols will have null probabilities assigned to them by the empirical probability estimator. In parallel, the derivative of the function $f(x) = -x\log(x)$ diverges near the origin. This characterizes a twofold impact, as on the one hand, the probabilities estimates of rare symbols is unreliable, since their expected number of occurrences is small, and they are often not observed in a sample. On the other hand, small errors in their probability estimations are the most problematic ones for entropy estimation, since they may accumulate to yield high errors in the final estimation, due to the high slope of the function $f$ in this region. This remarkable issue of the plug in estimator leads to disastrous results in the final estimation and this can only be reversed in the $N \gg S$ regime, where even the rarest symbols are expected to be observed multiple times. We illustrate this through the following example. 

\begin{example}
    Consider a Dirichlet distribution defined on the probability simplex of dimension $S=1000$, with parameters $\alpha_1 = \ldots = \alpha_S = 0.05$. From this distribution, we draw a probability vector $P$ whose entropy is 4.3722 nats. Following that, we draw a sample containing $N=1000$ instances from the distribution $P$, which we denote by $X^{1000}$. For the purpose entropy estimation, a sufficient statistic is the vector $h=(h_1, \ldots, h_{1000})$, with \begin{equation} \label{eq: profile}
        h_i = |\{j \ s.t \ n_j = i\}|, 
    \end{equation} where $|\cdot|$ stands for the cardinality of a set. Putting into words, $h_i$ is the number of symbols that were observed $i$ times in the sample. In Fig. \ref{fig: profile}, we present a discrete plot of the non-null portion of the vector $h$, which is commonly referred to as the profile of the sample, in the literature. 
    
    \begin{figure}[!htb]  
    \centering
    \includegraphics[width=\columnwidth]{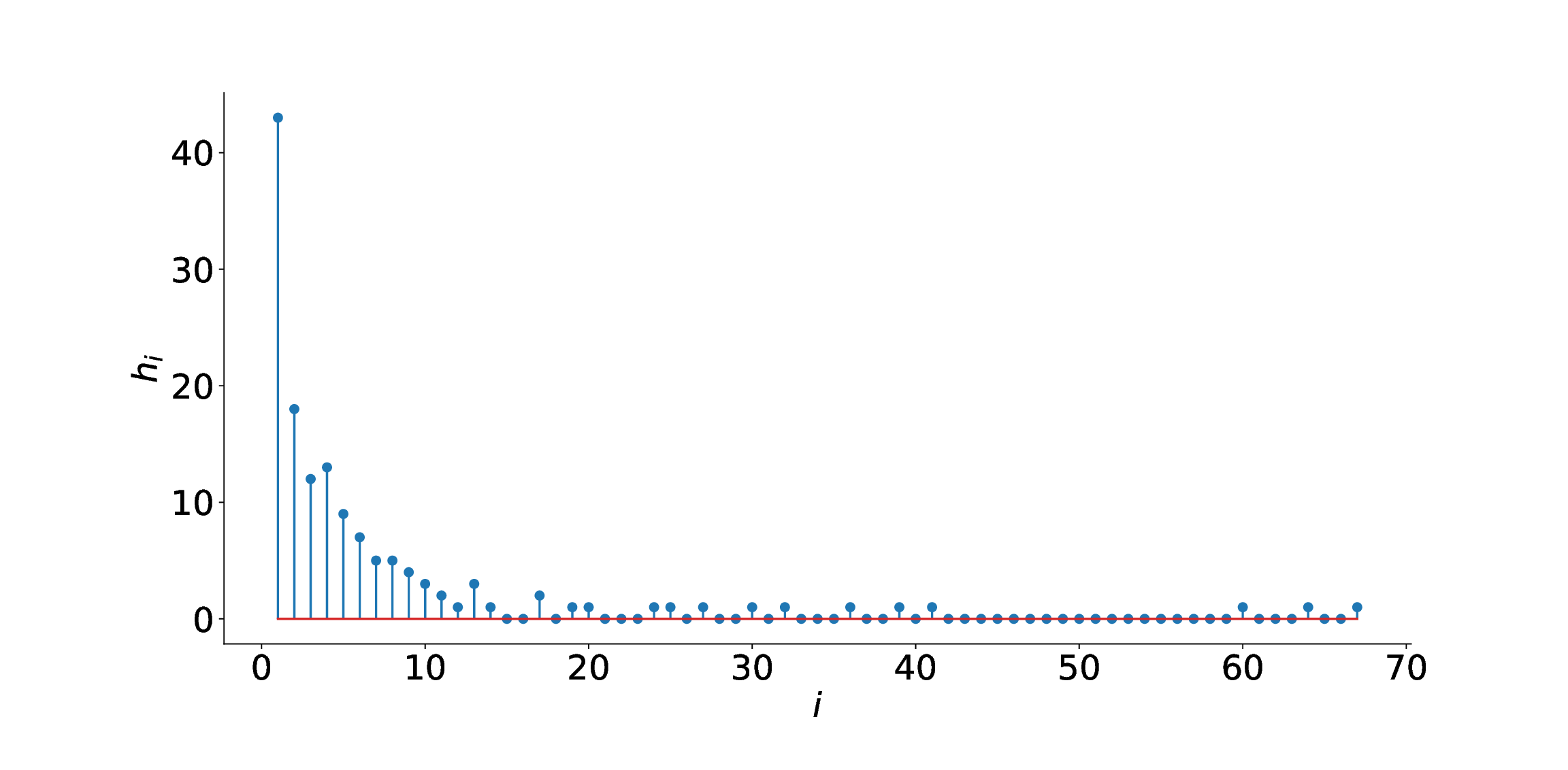}  
    \caption{Profile of the sample $X^{1000}$ with 1000 instances from $P$. The values of $h_i$ for $i>67$ were omitted from the plot since no symbol was observed more than 67 times.}
    \label{fig: profile}
\end{figure}

In this specific simulated sample, only 141 different outcomes were observed, out of a total 1000 possible ones, implying a total of 859 unseen symbols. The total probability mass of this set of unseen symbols, given by $m_0 = \sum \limits_{i \ s.t  \ n_i = 0} p_i$, is only 0.0467.

In the plug in estimator, the set of symbols that were observed $n$ times contributes to the final estimation error with \begin{equation}
    e_n = \sum \limits_{i \ s.t \ n_i=n} p_i\log(p_i) - \frac{n}{N} \log \left ( \frac{n}{N} \right). 
    \end{equation}

    In Fig. \ref{fig: erros}, $e_n$ is presented as a function of $n$.

    \begin{figure}[!htb]  
    \centering
    \includegraphics[width=\columnwidth]{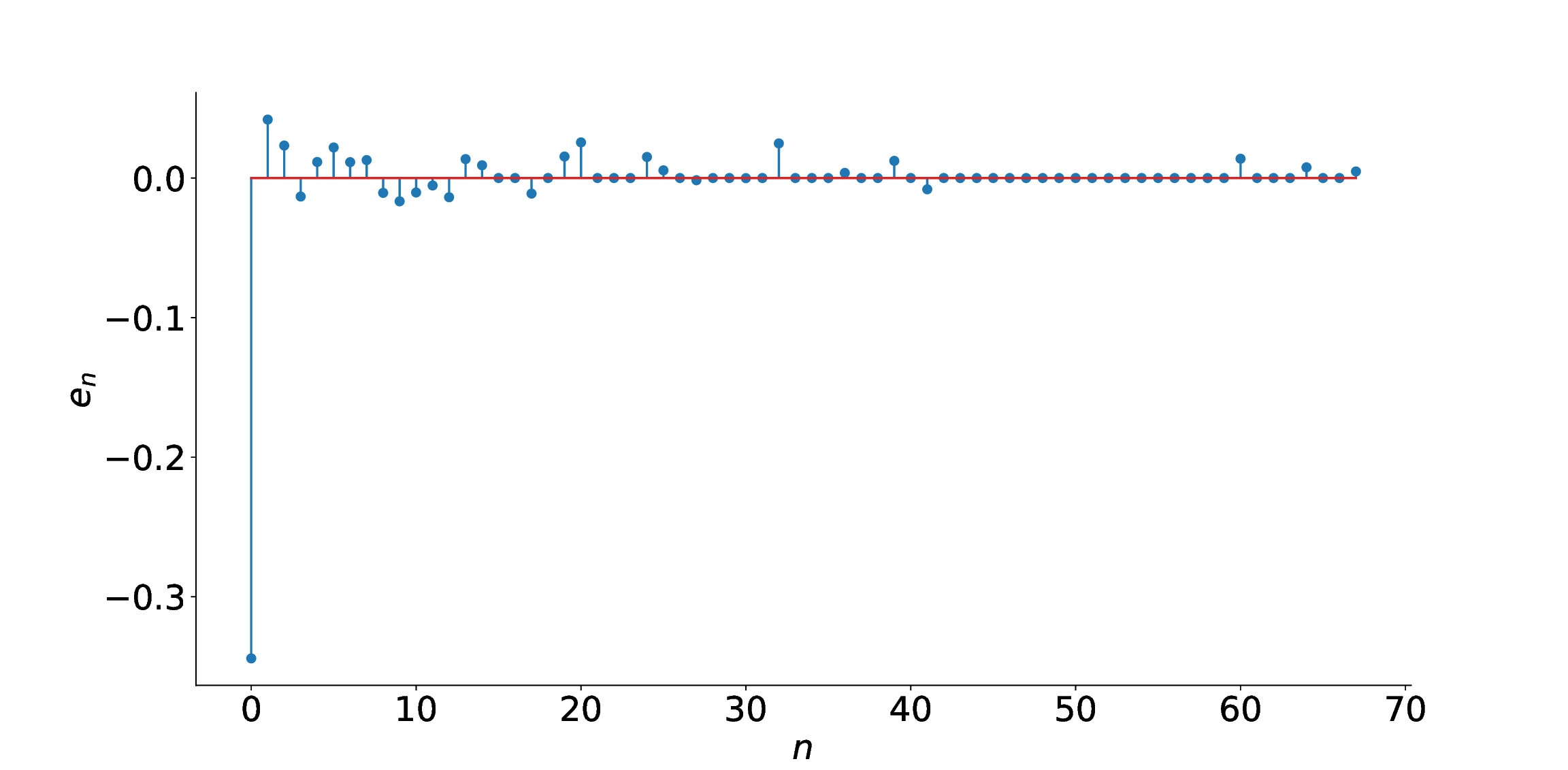}  
    \caption{Accumulated error in the plug estimator as a function of the number of occurrences, for the sample $X^{1000}$.}
    \label{fig: erros}
\end{figure}

    By taking a look at this this graphic, a striking observation can be made. A rare event, that only happens about 5\% of the time, is responsible for most of the negative bias incurred by the plug in approach. 

    This simple exemple puts into perspective the importance of compensating for the unseen symbols when estimating entropy in undersampled regimes. A great example of it is the Chao-Shen estimator, which achieves remarkable statistical properties by levaraging the Horvitz–Thompson estimator, that allows unbiased estimation of the population total of a given variable of interest across all possible outcomes in an experiment, when only a subset of them was observed \cite{chao2003}. In contrast, most of the other estimators, such as the Bayesian ones, will not explicitly take into account the unobserved symbols, unless the support size is known in advance.  In this paper, the contribution of the unseen symbols will be taken into account using tools that enables estimations of the number of unseen symbols and the total mass of these symbols. These tools will be introduced in the next Section.

\end{example}

\section{Estimating the properties of the unseen symbols} \label{sec: unseen}

In this Section we present existing estimators for two properties of the set of possible symbols that were not observed in a given sample, the total mass of this set and its cardinality.

\subsection{Estimating the total mass of the symbols that were observed $k$ times}

The total mass of the $h_k$ symbols that were observed $k$ times in a sample $X^N$ is given by \begin{equation} \label{eq: total mass}
    m_k = \sum \limits_{i \ s.t \ n_i = k} p_i.
\end{equation} 

A special case of it is the missing mass, $m_0$, which is the total mass of the unseen symbols, and its complement $1-m_0$ is the equivalent of the sample coverage defined in \cite{chao2003} for the Chao-Shen estimator. 

The most traditional approach to estimate $m_k$ is the Good-Turing estimator \cite{good1953}. This estimator was developed under a Poisson model where the sample size is modeled as a Poisson random variable with mean $N$ (in practice, $N$ is the actual sample size). This modeling choice makes the analysis easier, since under this assumption, $n_1, \ldots, n_S$ are independent Poisson random variables with means $\lambda_i = N p_i$, respectively, as opposed to a multinomial random variable, which would be the case with a fixed sample size.   

In \cite{lee2024}, the authors considered the dependencies among the variables $n_1, \ldots, n_S$, under the multinomial model, and derived an expression for the bias of the Good-Turing estimator, which is incorrectly identified as unbiased under the Poisson model. With the analysis, a minimal bias estimator for $m_k$ was proposed, as in \begin{equation} \label{eq: missing mass}
    \hat{m}_k = -\binom{N}{k} \sum \limits _{i=1}^{N-k} \frac{(-1)^i h_{k+1}}{\binom{N}{k+1}}.
\end{equation}

Furthermore, in that work, the authors developed a procedure to obtain a distribution dependent minimal mean squared error (MSE) estimator through a search problem.

\subsection{Estimating the number of unseen symbols}

In its most traditional formulation, the problem of estimating the number of unseen symbols is stated as follows. Given a sample $X^N$, with $N$ realizations of the target random variable, what is the number $U$ of symbols, not seen in $X^N$, that would be observed if a new sample $X^M$ were obtained. For convenience, we shall write $M = aN$, where $a$ is an amplification factor for the sample size. The most traditional estimator for $U$ was introduced in \cite{good1956} by Good and Toulmin. The Good-Toulmin estimator is unbiased for any $a$, but its variance grows indefinitely for $a>1$. In \cite{orlitsky2016}, the Good-Toulmin estimator was modified using a smoothing technique to obtain an optimal MSE estimation for any $a$ in the order of $\log(N)$, or smaller than that. Finally, in \cite{hao2020}, the smoothed Good-Toulmin estimator was extended to take into account the multiplicity. That is, given a sample $X^N$, how many symbols $U_{\mu}$ that were not observed in $X^N$ would be observed at least $\mu$ times in a new sample $X^M$? The estimator proposed in that work is given by \begin{equation}
    \label{eq: n_unseen}
    \hat{U}_{\mu} = \sum \limits _{i=1}^N s_i h_i,
\end{equation} where \begin{equation} \label{eq: s}
		s_i = - \sum \limits_{j=0}^{\min \{\mu - 1, i\}} (-a)^i (-1)^j \binom{i}{j}Pr(\text{Poi}(r) \geq i+j).
	\end{equation} Here $\text{Poi}(r)$ denotes a Poisson random variable with parameter $r$, which is a hyper-parameter that the authors keep fixed at \begin{equation} \label{eq: r}
	r = \frac{\log(n(a+1)^2/(a-1))}{2a}.
	\end{equation}

\section{Proposed Estimator} \label{sec: proposed}

Based on the argument presented in Section \ref{sec: empirical estimations}, we make the following observations: (i) the unseen symbols are responsible for most of the negative bias in entropy estimation, (ii) symbols that were observed with small frequencies are also problematic, since they have small probabilities, in the region of high slope of the function $f(x) = - x\log(f(x))$, and (iii) the empirical estimation of the probabilities of high frequency outcomes is more reliable for the purpose of entropy estimation.  

With that in mind, we conjecture that if the support is partitioned into these three subsets, it is possible to obtain reliable estimates if we use different strategies to estimate the contribution of each one of them separately, which includes taking into account the unseen symbols, even though the support size is unknown in advance. Therefore, we define the following subsets: 	\begin{equation} \label{eq: S1}
		S_1 = \{i \text{ s.t } n_i = 0\},
	\end{equation} 
	
	\begin{equation} \label{eq: S2}
		S_2 = \{i \text{ s.t } 0 < n_i \leq \lambda \},
	\end{equation}
	
	\begin{equation} \label{eq: S3}
		S_3 = \{i \text{ s.t } n_i > \lambda \}.
	\end{equation} 

    Putting into words, $S_1$ is the set of non-observed symbols, $S_2$ is the set of symbols that were rarely observed and $S_3$ is the set of symbols that were frequently observed. Here, the notion of rarity is associated with a threshold $\lambda$.

    We can then define the probability distribution $P_S = (P_{S_1}, P_{S_2}, P_{S_3})$, where $P_{S_i} = Pr(X \in S_i)$. With that, we may use the decomposability of the entropy --- see Section 2.5 of \cite{mackay2003} ---, which is a recursive property that allows us to write \begin{equation} \label{eq: decom}
\begin{split}
H(X) = H(P_S) 
  &+ P_{S_1} H(X \mid X \in S_1) \\
  &+ P_{S_2} H(X \mid X \in S_2)
  + P_{S_3} H(X \mid X \in S_3).
\end{split}
\end{equation} 

We propose to individually estimate every term of (\ref{eq: decom}), using an adequate strategy for each of them. 

\subsection{Estimation of the First Term}

To estimate the first term, one must determine the probability of each of the three subsets. For this, we use the total mass estimator, as in (\ref{eq: missing mass}), to obtain \begin{equation} \label{eq: ps1_est}
		\hat{P}_{S_1} = \hat{m}_{0},
	\end{equation} 
	
	\begin{equation} \label{eq: ps2_est}
		\hat{P}_{S_2} = \sum \limits_{i=1}^{\lambda} \hat{m}_i,
	\end{equation}
	
	\begin{equation} \label{eq: ps3_est}
		\hat{P}_{S_3} = 1 - \hat{P}_{S_1} - \hat{P}_{S_2}.
	\end{equation}

Equations (\ref{eq: ps1_est}), (\ref{eq: ps2_est}) and (\ref{eq: ps3_est}) can be replaced in the entropy definition, as in (\ref{eq: def}) to obtain a plug in estimation of the first term $\hat{H}(P_S)$

\subsection{Estimation of the Second Term}

The conditional entropy $H(X \mid X \in S_1)$ corresponds to the set of unseen symbols, which is responsible for most of the negative bias in the entropy estimation. To estimate $H(X \mid X \in S_1)$, we may use an estimate of the number of unseen symbols, as in (\ref{eq: n_unseen}),  with $\mu = 1$, and assume an uniform distribution in this set of rare symbols, for we do not have any other prior information available about their actual probabilities. With that, we estimate the conditional entropy as \begin{equation} \label{eq: cond1}
    \hat{H}(X \mid X \in S_1) = \log(\hat{U}_1).
\end{equation} 

Equations (\ref{eq: ps1_est}) and (\ref{eq: cond1}) can be combined to obtain an estimate $\hat{P}_{S_1}\hat{H}(X \mid X \in S_1)$ of the second term in (\ref{eq: decom}).

\subsection{Estimation of the Third Term}

The second conditional entropy $H(X \mid X \in S_2)$ is associated with symbols that are observed in the sample with small frequencies. These are also problematic, as their contributions to entropy estimation is sensitive to statistical fluctuations that increase the variance of the estimator. To mitigate that, we propose that these symbols are also considered uniformly distributed, and thus we obtain  \begin{equation} \label{eq: cond2}
    \hat{H}(X \mid X \in S_2) = \log(|S_2|).
\end{equation}

Combining (\ref{eq: ps2_est}) and (\ref{eq: ps3_est}), we obtain an estimate $\hat{P}_{S_2}\hat{H}(X \mid X \in S_2)$ for the third term of (\ref{eq: decom}).

\subsection{Estimation of the Fourth Term}

The last conditional entropy $H(X \mid X \in S_3)$ corresponds to the set of symbols that were observed in the sample with higher frequencies. Because these symbols have higher probabilities, the estimation of their contributions to entropy is less problematic. Therefore, we use the traditional Miller-Madow estimator \cite{miller1955} to obtain \begin{equation} \label{eq: cond3}
		\hat{H}(X \mid X \in S_3) = - \sum \limits_{i \ t.q \ n_i > \lambda} \frac{n_i}{N_{S_{3}}} \log \left ( \frac{n_{i}}{N_{S_3}} \right ) + \frac{|S_3|-1}{2N_{S_3}},
 	\end{equation} where $N_{S_3}$ is the amount of instances from $X^N$ that belongs to $S_3$, that is $N_{S_3} = \sum \limits_{i \ t.q \ n_i > \lambda} n_i$.

    Thus, we combine (\ref{eq: ps3_est}) and (\ref{eq: cond3}) to estimate the last term of (\ref{eq: decom}) as $\hat{P}_{S_3} \hat{H}(X \mid X \in S_3)$.
    
    Finally, we simply sum our estimates of the four terms to compose our proposed entropy estimator. 

\section{Experimental Evaluation} \label{sec: experiments}
In this Section, the proposed estimator will be experimentally tested and compared with 5 other estimators: plug in, Miller-Madow \cite{miller1955}, Chao-Shen \cite{chao2003}, Shrinkage \cite{hausser2009}, and the linear programming-based estimator in \cite{valiant2017}. For the experiments, the cardinality of the support will be fixed at $S=1000$, and different sampling scenarios will be considered, with $N$ varying in the set $\{100, 200, 300, 500, 1000, 2000, 5000 \}$. We carry the tests using the following four  distributions:

\begin{itemize}
    \item [(P1)] Uniform distribution ($H=6.9077$ nats),

    \item [(P2)] Randomly drawn distribution from a Dirichlet prior with parameters $\alpha_1 = \ldots = \alpha_S = 0.2$ ($H=5.5254$ nats),

    \item [(P3)] Randomly drawn distribution from a Dirichlet prior with parameters $\alpha_1 = \ldots = \alpha_S = 0.05$. ($H=4.4419$ nats),

    \item [(P4)] Randomly drawn distribution from a Dirichlet prior with parameters $\alpha_1 = \ldots = \alpha_S = 0.03$. ($H=3.8418$ nats),
\end{itemize}

The proposed estimator has two easily interpretable parameters. The first one is the threshold that separates the subsets $S_2$ and $S_3$, which we kept fixed at $\lambda = 3$. The other parameter is $a$, the amplification factor used to estimate the number of unseen symbols. The higher the value of $a$, the greater the chance that all non-observed symbols are included in this estimation. However, increasing $a$ also implies in an increase in the variance of the estimator of $U$, which in turns affects the performance of the overall estimator. To avoid this issue, we can rely on the intuition that if the missing mass is high, we need a bigger sample amplification in order to cover all the total probability mass of unobserved symbols, while a smaller $a$ should be used when the missing mass is already small. Using this intuition, we choose $a$ as a function of $\hat{m}_0$, which was already previously calculated in the estimation of the first term, as follows 

\begin{equation} \label{eq: a}
		a = \begin{cases}
			4 \times 10^5, \text{ if } \hat{m}_0 \geq 0.8, \\ 100, \text{ if } 0.7 \leq \hat{m}_0 < 0.8, \\ 8, \text{ if } 0.55 \leq \hat{m}_0 < 0.7, \\ 
			5, \text{ if } 0.4 \leq \hat{m}_0 < 0.55, \\ 2, \text{ if } 0.3 \leq \hat{m}_0 < 0.4, \\ 1.5, \text{ if } 0.15 \leq \hat{m}_0 < 0.3, \\ 1, \text{ if } \hat{m}_0 < 0.15.    
		\end{cases}
\end{equation} 

This look-up table built to select $a$ was empiracally fine tuned using only the ground-truth from the uniform distribution, and was kept fixed in every other experiment. In addition, the source code with the implementation of our estimator and additional experimental evaluation === to support the claim of generalizability of this approach --- with other distributions, with Zipf decaying laws and different support sizes, is available at  a public repository 
\footnote{ \url{https://github.com/gabrielfab0022/Partitioning-the-Sample-Space-for-a-More-Precise-Shannon-Entropy-Estimation}}.

To assess the quality of an entropy estimator, two statistical properties may be used. First, the bias of an estimator indicates how much the expected value of the estimator deviates from the true entropy, while the variance measures how much an estimator deviates around its own expected value. Minimizing these two measures is usually a conflicting object, which characterizes the well known bias-variance tradeoff. In general, the preferred evaluation metric is the MSE (or alternatively, the root mean squared error, RMSE), which is the sum of the squared bias and the variance. However, in some situations, minimizing the bias (or the variance) has a higher priority. For example, when the entropy is used as building block for further calculations, a lower bias is preferred than a lower variance, as systematic errors might propagate, potentially leading to systematically wrong results, while random fluctuations caused by a high variance may cancel out across many downstream uses.

\begin{figure}[!htb]
    \centering
    \subfloat[]{\includegraphics[width=64.3mm]{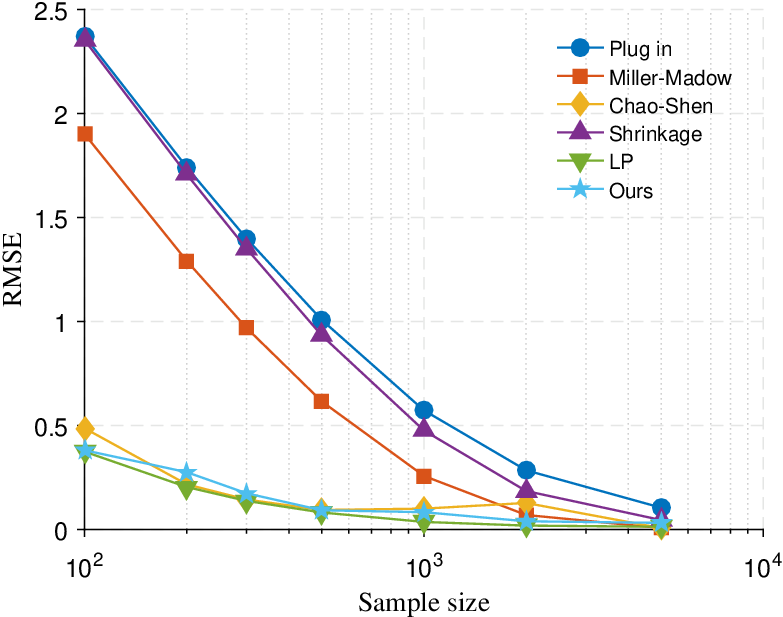}}\\[-1pt]
    \subfloat[]{\includegraphics[width=64.3mm]{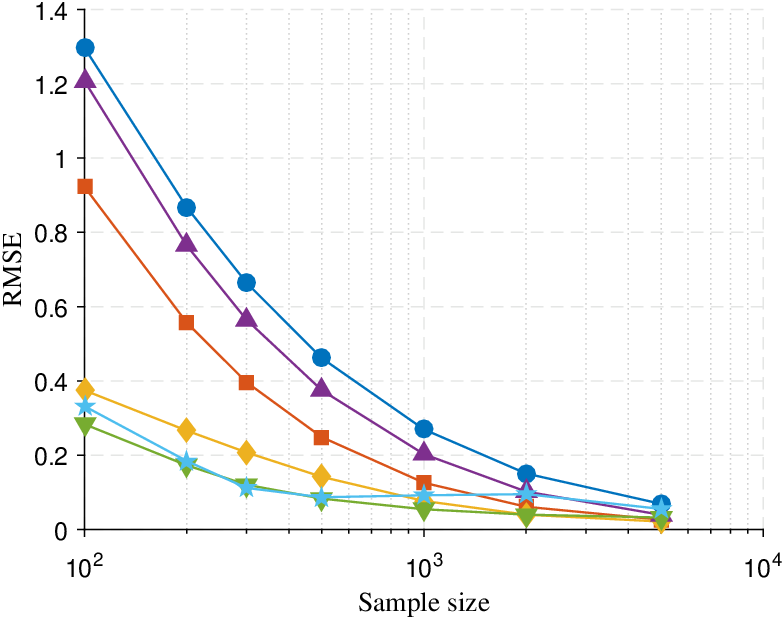}}\\[-1pt]
    \subfloat[]{\includegraphics[width=64.3mm]{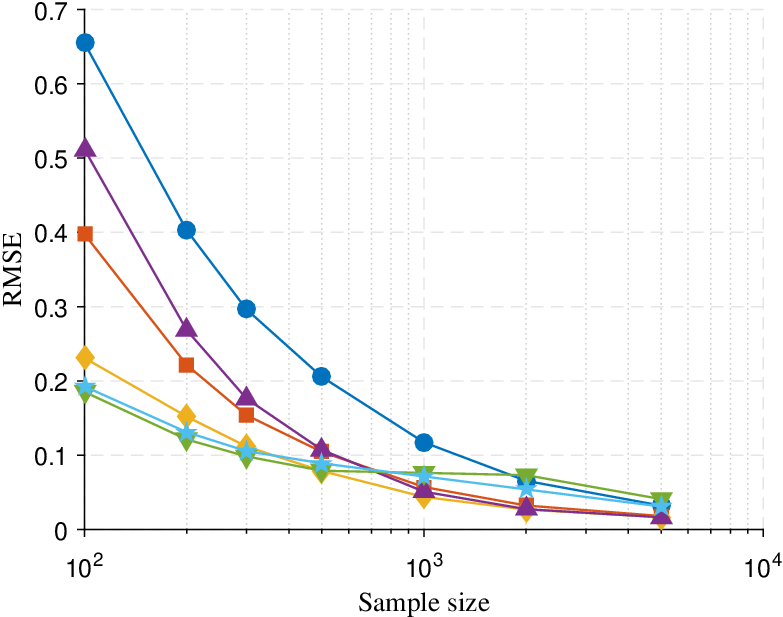}}\\[-1pt]
    \subfloat[]{\includegraphics[width=64.3mm]{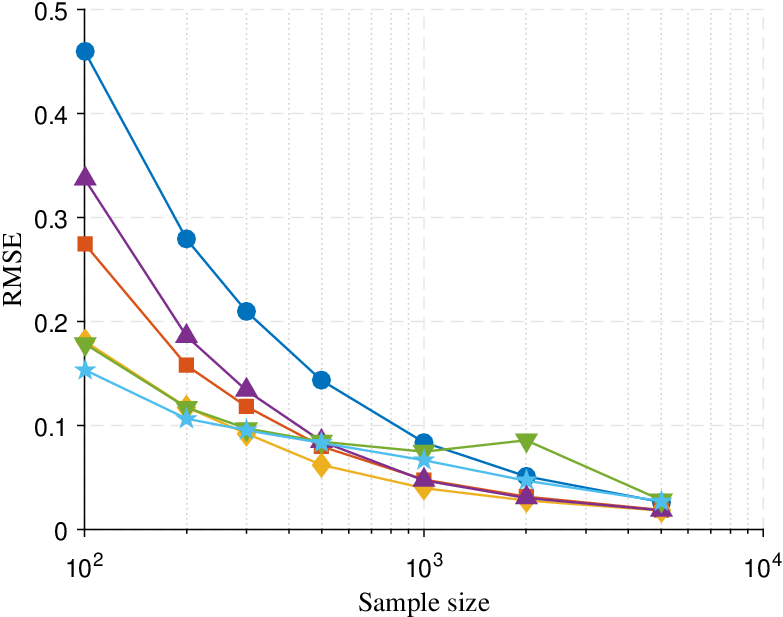}}
    \caption{RMSE comparison with distributions (a) P1, (b) P2, (c) P3, (d) P4}.
    \label{fig: rmse}
\end{figure}

\begin{figure}[!htb]
    \centering
    \subfloat[]{\includegraphics[width=64.3mm]{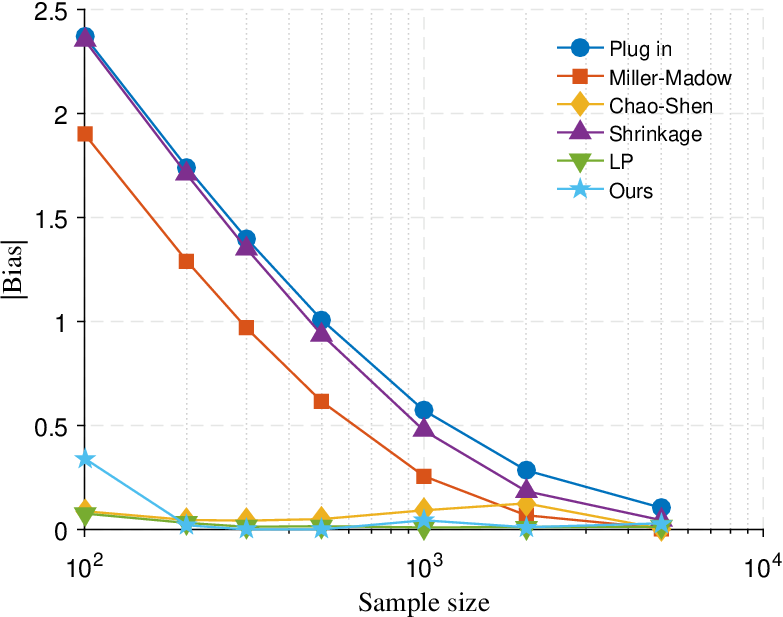}}\\[-1pt]
    \subfloat[]{\includegraphics[width=64.3mm]{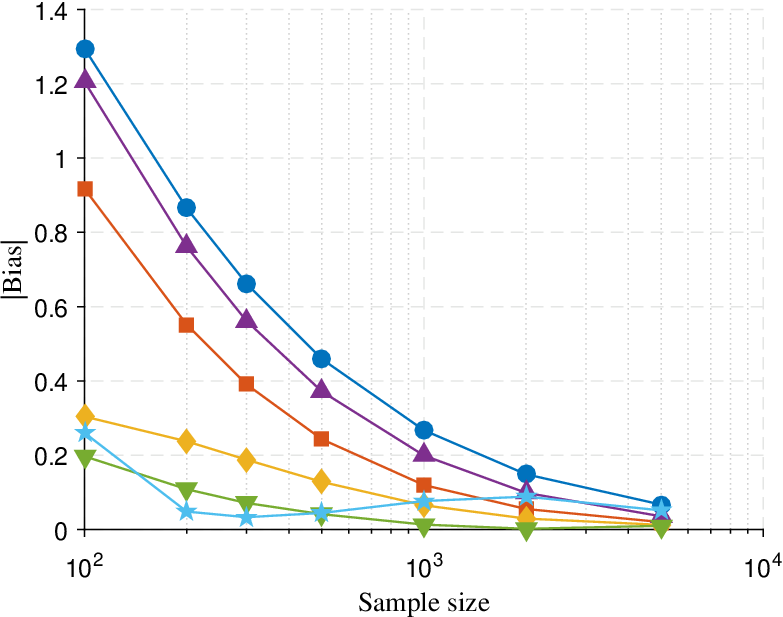}}\\[-1pt]
    \subfloat[]{\includegraphics[width=64.3mm]{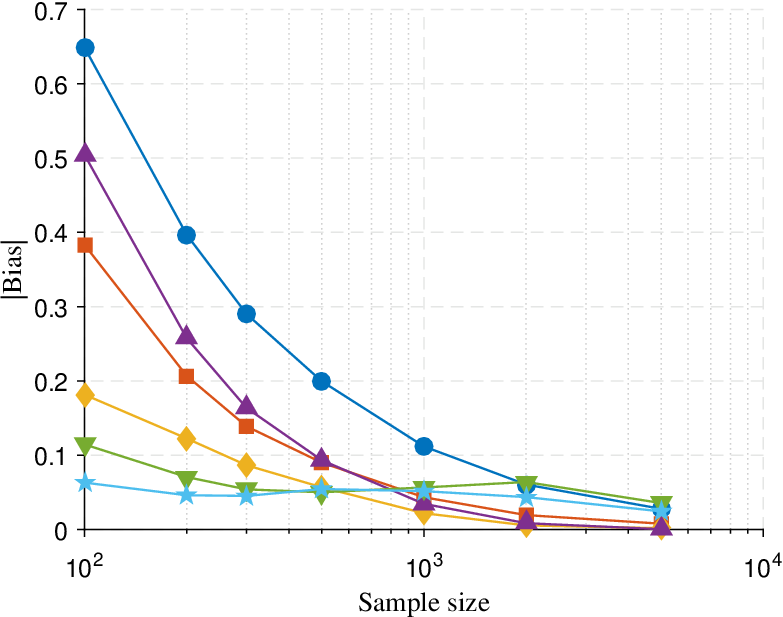}}\\[-1pt]
    \subfloat[]{\includegraphics[width=64.3mm]{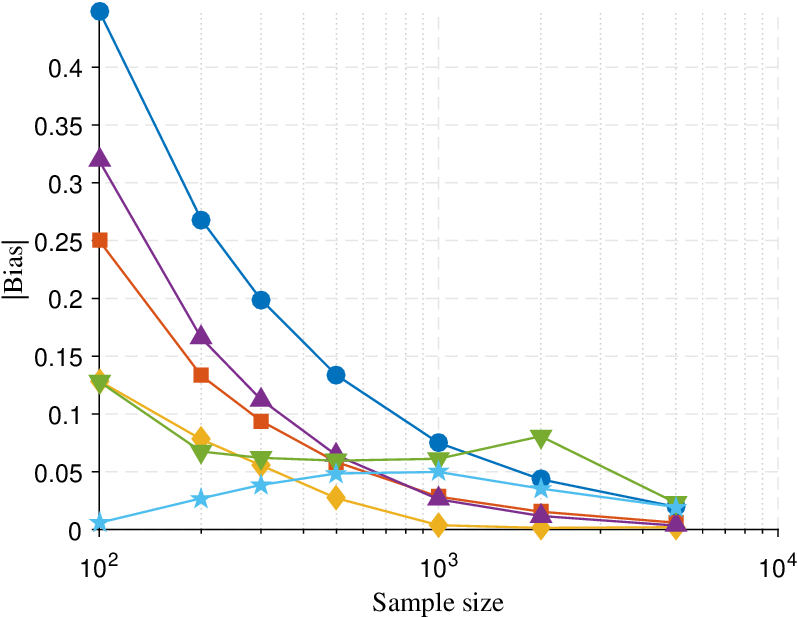}}
    \caption{Absolute value of the bias for distributions (a) P1, (b) P2, (c) P3, (d) P4}.
    \label{fig: bias}
\end{figure}

For this reason, we report both the RMSE and the bias of the proposed estimators, and of all the other five comparison methods, as a function of the sample size. To obtain these averages, 1000 independent simulations, with different samples, were carried for each combination of sample size and distribution. The results are presented in Figures \ref{fig: rmse} and \ref{fig: bias}.

As it can be seen in these graphics, the proposed estimator presents a comparable performance with those from the Chao-Shen estimator and from linear programming, both in terms of RMSE and bias. In highly undersampled regimes, when $N < S$, the performance of these three estimators is significantly superior to the Miller-Madow, the shrinkage and the plug in estimators, while all the estimators achieve low errors as the sample size increases.

\section{Conclusion} \label{sec: conclusions}

A new approach for entropy estimation of discrete sources was presented. The proposed approach explores the decomposability property of entropy, and uses estimators of the missing mass and of the number of unseen symbols to estimate the contributions of these symbols to the entropy. Experimental results with synthetic data sets show that this new estimator is able to achieve small RMSE and bias under small sample regimes, comparably to some of the best estimators available in the literature.  

We believe that the easy interpretability and the flexibility of the proposed estimator may allow for a lot of subsequent research. In particular, the accuracy of the proposed estimator could be further enhanced, for instance, by taking into account the multiplicity of the unseen symbols, as suggested in \cite{hao2020}, or by replacing the minimal bias estimator for the missing mass by the minimal MSE one as proposed in \cite{lee2024}. In future works, we also intend to further investigate how the parameters $a$ and $\lambda$ can be used to manage the tradeoff between estimator bias and variance. This study may also lead to a replacement of the look-up table used to set the parameter $a$  by a potentially more robust method. Moreover, the experimental evaluation, while promising, does not replace a rigorous theoretical understanding of the estimator's behavior, which should also be addressed in a future study.


\bibliographystyle{ieeetr}
\bibliography{refs}
\end{document}